\documentclass[letterpaper]{article} 
\usepackage{aaai23}  
\usepackage{times}  
\usepackage{helvet}  
\usepackage{courier}  
\usepackage[hyphens]{url}  
\usepackage{graphicx} 
\usepackage{natbib}  
\usepackage{caption} 
\usepackage{algorithm}
\usepackage{algorithmic}

\usepackage{newfloat}
\usepackage{listings}
\DeclareCaptionStyle{ruled}{labelfont=normalfont,labelsep=colon,strut=off} 
\floatstyle{ruled}
\newfloat{listing}{tb}{lst}{}
\floatname{listing}{Listing}
\usepackage{amssymb}

\usepackage{amsthm}
\newtheorem{definition}{Definition}
\newtheorem{example}{Example}

\usepackage{tikz}
\usepackage{tikz-qtree}

\newcommand{\twodots}{\mathrel{{.}\,{.}}\nobreak}

\usepackage{tabularx}
\usepackage{multirow}

\usepackage{tikz}
\usepackage{tikz-qtree}

\usepackage{array}
\newcolumntype{x}[1]{@{}>{\centering\arraybackslash\hspace{0pt}}p{#1}@{}}
\newcolumntype{y}[1]{@{}>{\raggedright\arraybackslash\hspace{0pt}}p{#1}@{}}

\title{\textsc{FastDiagP}: An Algorithm for Parallelized Direct Diagnosis}
\author {
    Viet-Man Le\textsuperscript{\rm 1},
    Cristian Vidal Silva\textsuperscript{\rm 2},
    Alexander Felfernig\textsuperscript{\rm 1},\\
    David Benavides\textsuperscript{\rm 3},
    Jos{\'e} Galindo\textsuperscript{\rm 3},
    Thi Ngoc Trang Tran\textsuperscript{\rm 1}
}
\affiliations {
    \textsuperscript{\rm 1} Graz University of Technology, Graz, Austria \\
    \textsuperscript{\rm 2} Universidad de Talca,  Talca, Chile\\
    \textsuperscript{\rm 3} University of Sevilla, Seville, Spain\\
    \{vietman.le,alexander.felfernig,ttrang\}@ist.tugraz.at,
    cvidal@utalca.cl, \{jagalindo,benavides\}@us.es
}

\begin{document}
\sloppy

\maketitle

\begin{abstract}
Constraint-based applications attempt to identify a solution that meets all defined user requirements. If the requirements are inconsistent with the underlying constraint set, algorithms that compute diagnoses for inconsistent constraints should be implemented to help users resolve the “no solution could be found” dilemma. \textsc{FastDiag} is a typical direct diagnosis algorithm that supports diagnosis calculation without predetermining conflicts. However, this approach faces runtime performance issues, especially when analyzing complex and large-scale knowledge bases. In this paper, we propose a novel algorithm, so-called \textsc{FastDiagP}, which is based on the idea of speculative programming. This algorithm extends \textsc{FastDiag} by integrating a parallelization mechanism that anticipates and pre-calculates consistency checks requested by \textsc{FastDiag}. This mechanism helps to provide consistency checks with fast answers and boosts the algorithm's runtime performance. The performance improvements of our proposed algorithm have been shown through empirical results using the \emph{Linux-2.6.3.33} configuration knowledge base.
\end{abstract}


\section{Introduction}\label{sec:intro}

In many applications of constraint-based representations such as \emph{knowledge-based configuration} \cite{Stumptner1997}, \emph{recommendation} \cite{FelfernigBurke2008}, \emph{automated analysis of feature models} \cite{Benavides2010}, and \emph{scheduling} \cite{Castillo2005}, there exist some scenarios where over-constrained formulations occur in the underlying constraint sets \cite{Felfernig2010_ECAI,Jannachetal2015}. Some examples thereof are inconsistencies between the knowledge base and a set of test cases \cite{Felfernig2014_Elsevier,le2021directdebug}, or inconsistencies between user requirements and the knowledge base \cite{felfernig2009plausible}. In such scenarios, \textit{diagnosis detection mechanisms} are essential to identify \textit{preferred minimal sets} of constraints (i.e., diagnoses) that are less important (from the user's point of view) and can be adapted or deleted to restore consistency in the knowledge base \cite{Reiter1987,Felfernig2012}.

\textit{Direct diagnosis} techniques have been recognized as efficient solutions in identifying faulty constraints without predetermining the corresponding conflict sets \cite{Felfernig2012, Felfernig2018}. \textsc{FastDiag} \cite{Felfernig2012} is a typical example of these techniques, designed to find \textit{one preferred minimal diagnosis} at a time within a given set of constraints ($C$). The algorithm divides $C$ into two subsets. If a subset is consistent, then diagnosis detection must not be applied to this subset since no diagnosis elements can be found in it. This way, $C$ can be reduced by half, and the algorithm returns one \textit{preferred minimal diagnosis} at a time. 
Although \textsc{FastDiag} works efficiently in many scenarios, there exist cases where it faces runtime issues, especially in \textit{interactive settings}, where users are interacting with a configurator with a huge and complex knowledge base and expect to receive instant responses \cite{Felfernig2018}.

Consistency checking is an expensive computational task that makes up most of \textsc{FastDiag}'s execution time \cite{Felfernig2014}. 
A practical solution for this issue is to pre-calculate in parallel consistency checks potentially required by \textsc{FastDiag}. This solution provides fast answers for consistency checks (via simple lookup in a list of already-calculated consistency checks instead of a direct solver call), which helps to accelerate the algorithm's execution. Based on this idea, we propose in this paper a novel diagnosis detection algorithm, so-called \textsc{FastDiagP}, dealing with \textsc{FastDiag}'s run-time limitation. \textsc{FastDiagP} is a parallelized version of \textsc{FastDiag}, adopting the \emph{speculative programming principle} \cite{Burton1985} to pre-calculate consistency checks. Although this principle is not new, modern CPUs with integrated parallel computation capabilities now make it possible to implement some speculative approaches.

The contributions of our paper are \emph{three-fold}. \emph{First}, we show how to parallelize direct diagnosis based on a flexible \textit{look-ahead strategy} to scale its performance depending on the number of available computing cores. \emph{Second}, we show how to integrate the proposed approach into \textsc{FastDiag}, which is applicable to interactive constraint-based applications. \emph{Third}, using the inconsistent \emph{Linux-2.6.33.3} feature model taken from Diverso Lab's benchmark \cite{Heradio2022}, we show the performance improvements of \textsc{FastDiagP} when working with large-scale configuration knowledge bases. Particularly, this algorithm improves the performance of diagnosis detection tasks scaling with available CPU cores, making it possible to efficiently solve more complex diagnosis problems. 


\section{Related Work}\label{sec:related-work}

\emph{Solution Search}. The increasing size and complexity of knowledge bases have led to the need of improving search solutions \cite{Bordeauxetal2009,Gentetal2018}. Such solutions have been implemented as parallelization algorithms in different contexts. For example, \cite{Bordeauxetal2009} propose a parallelization approach 
to determine solutions for sub-problems independent of available cores. Due to the development of multi-core CPU architectures, parallelization approaches have become increasingly popular to exploit computing resources better and obtain expected results more efficiently. 


\emph{Conflict Detection}. Determining more efficiently minimal conflicts is a core requirement in many application settings \cite{Jannachetal2015}. In constraint-based reasoning scenarios, the \textsc{QuickXplan} algorithm \cite{junker2004quickxplain} is applied to identify minimal conflict sets following the divide-and-conquer-based strategy. Although this algorithm helps to reduce the number of needed consistency checks significantly and supports interactive settings, it faces runtime performance issues. In this context, \cite{Vidal2021} proposes a conflict detection approach based on speculative programming, called the so-called {\textsc{parallelized QuickXPlain}. The empirical results show that this approach helps to significantly improve the runtime performance of \textsc{QuickXplan}.

\emph{Conflict Resolution}. Conflict detection is the basis of \emph{conflict resolution} that attempts to identify sets of minimal diagnoses  \cite{Reiter1987,MarquesSilva2013}. For instance,  \cite{Jannachetal2015,Jannachetal2016} propose approaches to parallelize the computation of hitting sets (diagnoses). In these studies, a level-wise expansion of a breadth-first search tree is adopted to parallelize model-based diagnosis \cite{Reiter1987} and compute minimal cardinality diagnoses. However, the determination of individual conflict sets is still a sequential process (based on \textsc{QuickXPlain} \cite{junker2004quickxplain}). In another study, \cite{Jannachetal2016} replace the level-wise expansion with a full hitting set parallelization and take into account additional mechanisms to ensure diagnosis minimality. Although the mentioned approaches focus on the parallelization of conflict resolution, they do not offer solutions to increase the efficiency of conflict detection. In this paper, based on the speculative programming principle \cite{Burton1985}, we propose an algorithm integrating a parallelized conflict resolution mechanism that helps to significantly improve the runtime performance of direct diagnosis processes.


\section{Example Configuration Knowledge Base}\label{sec:working-example}

For demonstration purposes, we introduce a working example with a configuration knowledge base from the \textit{smartwatch} domain. A \emph{Smartwatch} must have at least one type of \emph{Connector} and \emph{Screen}. The connector can be one or more out of the following: \emph{GPS}, \emph{Cellular}, \emph{Wifi}, or \emph{Bluetooth}. The screen can be either \emph{Analog}, \emph{High Resolution}, or \emph{E-ink}. A \textit{Smartwatch} may include a \emph{Camera} and a \emph{Compass}. Besides, \emph{Compass} requires \emph{GPS} and \emph{Camera} requires \emph{High Resolution}. Finally, \emph{Cellular} and \emph{Analog} exclude each other.


Our simpliﬁed conﬁguration knowledge base can be represented as a \emph{conﬁguration task} which is deﬁned as a \emph{constraint satisfaction problem} (CSP) \cite{rossi2006handbook}. A \emph{conﬁguration task} and its \emph{configuration} (solution) are deﬁned as follows \cite{Hotz2014}:

\begin{definition}[Configuration task]
\label{def:configurationtask}
    \emph{A configuration task can be defined as a CSP $(V, D, C)$ where $V=\{v_1, v_2 \twodots v_n\}$ is a set of variables, $D=\{dom(v_1), dom(v_2) \twodots dom(v_n)\}$ is a set of domains for each of the variables in $V$, and $C = C_{KB} \cup C_R$ is a set of constraints restricting possible solutions for a configuration task. $C_{KB}$ represents the configuration knowledge base (the configuration model) and $C_R$ represents a set of user requirements.}
\end{definition}

\begin{definition}[Configuration]
\label{def:configuration}
    \emph{A configuration (solution) $S$ for a given configuration task $(V, D, C)$ is an \emph{assignment} $A=\{ v_1=a_1 \twodots v_n=a_n \}$, $a_i \in dom(v_i)$. $S$ is \textbf{valid} if it is \emph{complete} (i.e., each variable in $V$ has a value) and \emph{consistent} (i.e., $S$ fulfills the constraints in $C$).}
\end{definition}

\begin{example}[CSP-based representation of a Smartwatch configuration task]
\label{ex:csp}
    \emph{A CSP-based representation of a conﬁguration task $(V,D,C=C_{KB} \cup C_R)$ that can be generated from our simplified configuration knowledge base is the following (see \textit{Table \ref{table:ex_kb}} for constraints in $C_{KB}$ and $C_R$):}
    
    \begin{itemize}
        \item $V=\{Smartwatch$, $Connector$, $Screen$, $Camera$, $Compass$, $GPS$, $Cellular$, $Wifi$, $Bluetooth$, $Analog$, $High$ $Resolution$, $E$-$ink\}$,
        \item $D=\{dom(Smartwatch) \twodots dom(E$-$ink)\}$,  where $dom(v_i)$ $=\{(t)rue,(f)alse\}$,
        \item $C_{KB}=\{c_0 \twodots c_9$\}, $C_R=\{c_{10} \twodots c_{13}\}$.
    \end{itemize}
    
    \emph{According to \textit{Table \ref{table:ex_kb}}, we can observe that some constraints in $C_R$ are inconsistent with the constraints in $C_{KB}$. For instance, $c_{10}$ and $c_{11}$ in $C_R$ are inconsistent with $c_9$ in $C_{KB}$. Therefore, no solution can be found for this conﬁguration task. For related faulty constraints, see \textit{Example \ref{ex:conflict}}.}\qed
\end{example}

\begin{table}[!t]
\small
    \begin{center}
    \begin{tabular}{x{0.6cm}y{4cm}x{0.7cm}y{1.9cm}}
    \hline
          & \multicolumn{3}{l}{\textbf{CSP representation}}\tabularnewline
    \hline
          & \multicolumn{3}{l}{\textbf{Constraints in the knowledge base} - $C_{KB}$}\tabularnewline
    $c_0$ & \multicolumn{3}{l}{$Smartwatch=t$}\tabularnewline
    $c_1$ & \multicolumn{3}{l}{$Smartwatch \leftrightarrow Connector$} \tabularnewline
    $c_2$ & \multicolumn{3}{l}{$Smartwatch \leftrightarrow Screen$} \tabularnewline
    $c_3$ & \multicolumn{3}{l}{$Camera \rightarrow Smartwatch$} \tabularnewline
    $c_4$ & \multicolumn{3}{l}{$Compass \rightarrow Smartwatch$} \tabularnewline
    $c_5$ & \multicolumn{3}{l}{$Connector \leftrightarrow (GPS \vee Cellular \vee Wifi \vee Bluetooth)$} \tabularnewline
    $c_6$ & \multicolumn{3}{l}{$Screen \leftrightarrow xor(Analog, High$ $Resolution, E$-$ink)$} \tabularnewline
    $c_7$ & \multicolumn{3}{l}{$Camera \rightarrow High$ $Resolution$} \tabularnewline
    $c_8$ & \multicolumn{3}{l}{$Compass \rightarrow GPS$} \tabularnewline
    $c_9$ & \multicolumn{3}{l}{$\neg(Cellular \wedge Analog)$} \tabularnewline
    \hline
    & \multicolumn{3}{l}{\textbf{User requirements} - $C_R$}\tabularnewline
    $c_{10}$ & \hspace{6pt}$Cellular=t$ & $c_{12}$ & $Compass=t$\tabularnewline
    $c_{11}$ & \hspace{6pt}$Analog=t$ & $c_{13}$ & $GPS=f$\tabularnewline
    \hline
    \end{tabular}
    {\caption{Constraints in $C_{KB} = \{c_0 \twodots c_9\}$ derived from our simplified configuration knowledge base, $C_R = \{c_{10} \twodots c_{13}\}$ is a set of user requirements.}

    \label{table:ex_kb}}
    \end{center}
\end{table}

Due to inconsistent constraints in the knowledge base/user requirements, the reasoning engine (e.g., constraint solver) cannot determine a solution. In this context, identifying explanations (in terms of diagnoses) is extremely important to help users adapt their requirements and thus restore consistency. In the next section, we introduce basic concepts regarding diagnoses and preferred minimal diagnoses. Also, we revisit the \textsc{FastDiag} algorithm \cite{Felfernig2012} and show how a preferred minimal diagnosis can be determined using this algorithm.

\section{Determination of Preferred Diagnoses }\label{sec:determining-diagnoses}

Since the notions of a (minimal) \emph{conflict} and a (minimal) \emph{diagnosis} will be used in the following sections, we provide the corresponding definitions here. We use \emph{consistent}($C$) to denote that the constraint set $C$ is consistent, and \emph{inconsistent}($C$) to denote that the constraint set $C$ is inconsistent.

A \emph{conflict set} can be defined as a minimal set of constraints that is responsible for an inconsistency, i.e., a situation in which no solution can be found for a given set of constraints $C$ (see \textit{Definition \ref{def:conflict-set}}).

\begin{definition}[Conflict set]
\label{def:conflict-set}
    \emph{A \emph{conﬂict set} is a set $CS \subseteq C:$ \emph{inconsistent}($CS$). $CS$ is \emph{minimal} iff $\nexists CS': CS' \subset CS$.}
\end{definition}

\begin{example}[Minimal conflict sets]
\label{ex:conflict}
\emph{We are able to identify the following minimal conflict sets: $CS_1=\{c_{10},c_{11}\}$ and $CS_2=\{c_{12},c_{13}\}$. The minimality property is fulﬁlled since $\nexists CS_3: CS_3 \subset CS_1$ and $\nexists CS_4: CS_4 \subset CS_2$.}\qed
\end{example}

In order to resolve all conflicts, we need to determine corresponding hitting sets (also denoted as diagnoses \cite{Reiter1987}) that have to be adapted or deleted to make the user requirements consistent with the knowledge base. Based on the definition of a conflict set, we now introduce the definition of a \textit{diagnosis task} and a corresponding \textit{diagnosis}.

\begin{definition}[Diagnosis task]
\label{def:diag-task}
\emph{A diagnosis task can be defined by a tuple $(C_R, C_{KB})$, where $C_R$ is a set of user requirements to be analyzed and $C_{KB}$ is a set of constraints specifying the configuration knowledge base.}
\end{definition}

\begin{definition}[Diagnosis and Maximal Satisfiable Subset]
\label{def:diagnosis}
\emph{A \emph{diagnosis} $\Delta$ of a diagnosis task $(C_R, C_{KB})$ is a set $\Delta \subseteq C_R:$ \emph{consistent}($C_R \setminus \Delta \cup C_{KB}$). $\Delta$ is \emph{minimal} iff $\nexists \Delta' : \Delta' \subset \Delta$. A complement of $\Delta$ (i.e., $C_R \setminus \Delta$) is denoted as \emph{Maximal Satisfiable Subset} (MSS) $\Omega$.}
\end{definition}

\begin{example}[Minimal diagnoses]
\label{ex:diagnosis}
\emph{Applying conflict-directed diagnosis approaches \cite{Reiter1987} to the diagnosis task $(C_R=\{c_{10} \twodots c_{13}\}, C_{KB}= \{c_0 \twodots c_9\})$ presented in \emph{Examples \ref{ex:csp}} and \emph{\ref{ex:conflict}}, the corresponding minimal diagnoses are the following: $\Delta_1 = \{ c_{10}, c_{12} \}$, $\Delta_2 = \{ c_{10}, c_{13} \}$, $\Delta_3 = \{ c_{11}, c_{12} \}$, and $\Delta_4 = \{ c_{11}, c_{13} \}$.}\qed
\end{example}

\subsection{Preferred Diagnosis}
\label{sec:preferred-diagnosis}

To resolve given inconsistencies, a user has to choose a diagnosis consisting the constraints that need to be adapted/deleted. In this context, a diagnosis less important to the user is chosen first \cite{junker2004quickxplain}. Such a diagnosis is a so-called \textit{``preferred diagnosis''} \cite{marquessilva2014preferred} (defined in \textit{Definition \ref{def:preferred-diag}} based on \textit{Definitions \ref{def:total-order}} 
and \textit{\ref{def:apreference}}).

\begin{definition}[Strict total order]
\label{def:total-order}
\emph{Let $<$ be a \emph{strict total order} over the constraints in $C = \{c_1 \twodots c_m\}$ which is represented as $\langle c_1 < c_2 < \twodots < c_m \rangle$, i.e., $c_i$ is preferred over $c_{i+1}$.}
\end{definition}


\begin{definition}[Anti-lexicographic preference, A-Preference]
\label{def:apreference}
\emph{Given a \emph{strict total order} $<$ over $C$, a set $X \subseteq C$ is \emph{anti-lexicographically} preferred over another set $Y \subseteq C$ (denoted $X <_{antilex} Y$) iff $\exists_{i \leq k \leq m}:c_k \in Y \setminus X$ and $X \cap \{c_{k+1} \twodots c_m \} = Y \cap \{c_{k+1} \twodots c_m \}$.}
\end{definition}

\begin{definition}[Preferred diagnosis]
\label{def:preferred-diag}
\emph{A minimal diagnosis $\Delta$ for a given diagnosis task $(C_R, C_{KB})$ is a preferred diagnosis for $(C_R, C_{KB})$ iff $\forall \Delta': \Delta' <_{antilex} \Delta$.}
\end{definition}

Given a strict total order $<$ over a set of constraints, there exists a unique preferred diagnosis.

\begin{example}[A preferred diagnosis]
\label{ex:preferred-diagnosis}
\emph{Given two minimal diagnoses $\Delta_3=\{c_{11},c_{12}\}$, $\Delta_4=\{c_{11},c_{13}\}$ and the strict total order $\langle c_{11} < c_{12} < c_{13} \rangle$, we can say:}
\begin{itemize}
    \item \emph{$\Delta_3$ is \emph{anti-lexicographically} preferred over $\Delta_4$ since $c_{13} \in \Delta_4 \setminus \Delta_3$ with $\Delta_4 \cap \emptyset$ = $\Delta_3 \cap \emptyset$.}
    \item \emph{$\Delta_4$ is a preferred diagnosis since it contains $c_{13}$ that is less important than $c_{12}$ presented in $\Delta_3$}.\qed
\end{itemize}
\end{example}

\subsection{\textsc{FastDiag}}

\textsc{FastDiag} \cite{Felfernig2012} determines a diagnosis without the need of conflict detection and a related derivation of hitting sets \cite{Reiter1987}. \textit{Algorithms \ref{alg:fastdiag}} and \textit{\ref{alg:fd}} below show a variant of \textsc{FastDiag}, where \textit{Algorithm \ref{alg:fd}} - \textsc{FD} determines a \textit{maximal satisfiable subset} $\Omega$ instead of a \textit{minimal correction subset} as in the original version presented in \cite{Felfernig2012}.

\begin{algorithm}[b]
\caption{\textsc{FastDiag}$(C, B): \Delta$}
\small
\begin{algorithmic}[1]\label{alg:fastdiag}
\IF{$C=\emptyset$ or \textsc{Consistent}$(B \cup C)$}
\STATE \emph{return}($\emptyset$)
\ELSE
\STATE \emph{return}($C \setminus \textsc{FD}(C,B,\emptyset)$)
\ENDIF
\end{algorithmic}
\end{algorithm}

\textit{Algorithm \ref{alg:fastdiag}} - \textsc{FastDiag} includes two variables $C$ and $B$, where $C$ consists of potentially faulty constraints in $C_R$ and $B$ contains correct constraints in $C_{KB}$. The constraint ordering in $C$ conforms to the definition of the \textit{strict total order} (see \textit{Definition \ref{def:total-order}}). If \emph{inconsistent}($B \cup C$), then \textit{Algorithm \ref{alg:fd}} - \textsc{FD} is activated to identify constraints in $C$ that are responsible for the inconsistency. \textsc{FD} determines an MSS $\Omega$, from which the corresponding minimal diagnosis can be derived ($\Delta=C \setminus \Omega$). In \textsc{FD}, if \textit{consistent}($B \cup C$), then $C$ is returned since no diagnosis elements can be found and $C$ becomes part of the MSS. If there is only one constraint $c_i$ in $C$, then $c_i$ is an element of a conflict since \emph{inconsistent}($B \cup C$). This element is removed from $C$ (by returning an empty set) to guarantee that $C$ is an \textsc{MSS}. If \emph{inconsistent}($B \cup C$) and $C$ has more than one element, the \textsc{Split} function is called to divide $C$ into two subsets $C_l = \{ c_1 \twodots c_{k} \}$ and $C_r =\{ c_{k+1} \twodots c_{n} \}$, where $k=\lfloor\frac{n}{2}\rfloor$. Finally, \textsc{FastDiag} returns an MSS and the corresponding minimal diagnosis $\Delta$. 

\begin{algorithm}[t]
\caption{$\textsc{FD}(C=\{c_1 \twodots c_n\},B,\rho): \Omega$}
\small
\begin{algorithmic}[1] \label{alg:fd}
\IF{$\rho \neq \emptyset$ and $\textsc{Consistent}(B \cup C)$}
\STATE \emph{return}($C$)
\ENDIF
\IF{$|C|=1$}
\STATE \emph{return}($\emptyset$)
\ENDIF
\STATE \textsc{Split}$(C, C_l, C_r)$
\STATE $\Omega_2 \leftarrow \textsc{FD}(C_l,B,C_r)$
\STATE $\Omega_1 \leftarrow \textsc{FD}(C_r,B \cup \Omega_2,C_l \setminus \Omega_2)$
\STATE \emph{return}($\Omega_1 \cup \Omega_2$)
\end{algorithmic}
\end{algorithm}


The parameter $\rho$ in the FD algorithm plays an important role in avoiding redundant consistency checks. Assigning $C_r$ to $\rho$ (see \textit{line 8}) triggers a consistency check for $B \cup C_l$. If \emph{consistent}($B \cup C_l$), $C_l$ will be returned by the \textsc{FD} call at \textit{line 2}. The \textsc{FD} call at \textit{line 9} does not trigger a consistency check in \textit{line 1} since $C_l \setminus \Omega_2 = \emptyset$, i.e., $B \cup C_r \cup \Omega_2$ = $B \cup C_r \cup C_l$ = $B \cup C$ has been already checked. The details of how \textsc{FastDiag} works are shown in \textit{Figure \ref{fig:fd-ex-tree}} on the basis of our working example.

\begin{figure}[ht]
\center
\centerline{\includegraphics[scale=0.69]{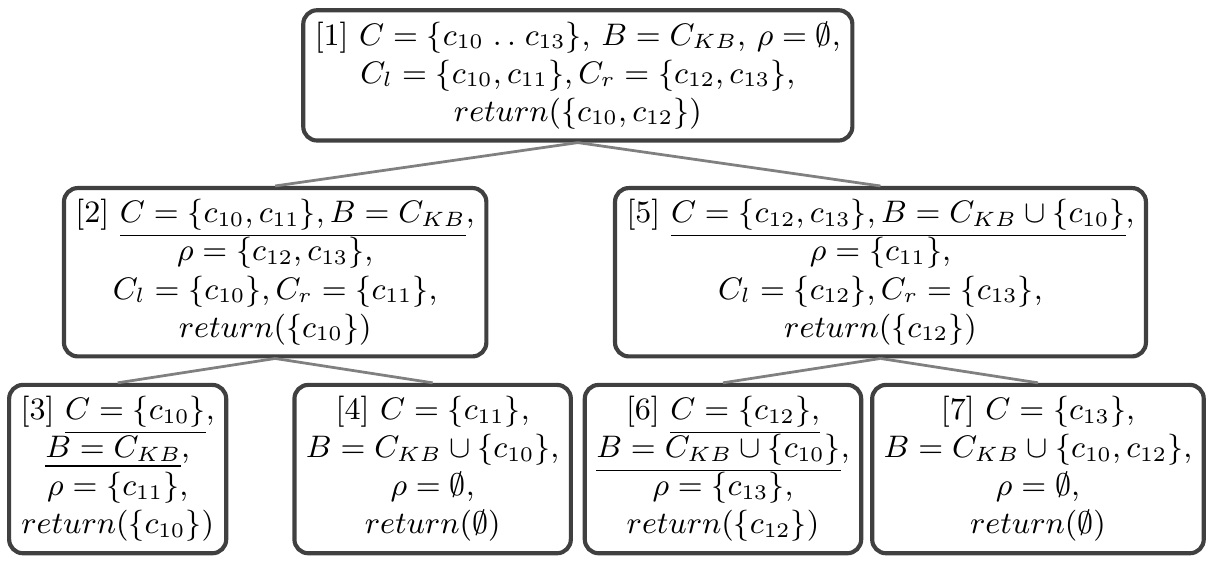}}
\caption{\textsc{FD} execution trace for $C=\{c_{10} \twodots c_{13}\}$ and $B=C_{KB}$. \textsc{FD} determines a MSS $\Omega =\{ c_{10},c_{12} \}$ and the corresponding diagnosis $\Delta=\{ c_{11},c_{13} \}$. The underlined $C$ and $B$ instances denote \textsc{FD} consistency checks.}
\label{fig:fd-ex-tree} 
\end{figure}

\section{\textsc{FastDiagP} - Parallelized \textsc{FastDiag}}\label{sec:fastdiagp}

\textbf{General idea.} \textsc{FastDiagP} is the parallelized version of \textsc{FastDiag}, in which we integrate a \emph{look-ahead} mechanism adopting the speculative programming principle \cite{Burton1985} into the \textsc{Consistent} function. The \emph{look-ahead} mechanism performs two tasks: (1) anticipating potential consistency checks that \textsc{FastDiag} might need in the near future, and (2) scheduling the asynchronous execution of anticipated consistency checks.

To ensure correct and useful anticipated consistency checks (i.e., \textsc{FD} will request consistency checks' results in its next calls), the anticipation of the \emph{look-ahead} mechanism complies with the two following principles \emph{P1} and \emph{P2}:

- \emph{P1 (Following two assumptions concerning the consistency of $B \cup C$)}: In each recursive step of \textsc{FD}, the decision for the next recursive call depends on the consistency of the current $B \cup C$. If \emph{inconsistent}($B \cup C$), \textsc{FD} applies the \emph{divide-and-conquer} strategy to $C$. Otherwise, a strategy for $\rho$ is used, holding the sibling half of $C$. Thus, an \emph{inconsistency} assumption helps to discover the next level of the \textsc{FD} execution tree, while a \emph{consistency} one helps to exploit the sibling of the current call. This way, the \emph{look-ahead} mechanism can generate all needed consistency checks without redundancy.
    
- \emph{P2 (Complying with the \emph{divide-and-conquer} strategy of \textsc{FastDiag})}: In each recursive step of \textsc{FD}, when \emph{inconsistent}($B \cup C$) and $C$ is not a singleton, consistency checks for the two halves of $C$ are triggered by \textsc{FD}. Thus, regarding the \emph{look-ahead} mechanism, when the current consideration set is not a singleton, the \emph{divide-and-conquer} strategy is applied to both \emph{consistency} and \emph{inconsistency} assumption branches to obtain the same effect as \textsc{FD} can make. The current consideration set could be $C$ or $\rho$.

Besides, the anticipation considers the computer resources in terms of available CPU cores ($\#cores$) in order to generate adequate consistency checks. For instance, the current \textsc{FD} execution needs the consistency check for $C= \{ c_1,c_2 \}$. The algorithm also knows that $B=\emptyset$, $\rho=\{ c_3,c_4 \}$, i.e., the remaining constraints will be checked if $C$ is consistent, and the system has a 4-cores CPU. In this context, the \emph{look-ahead} mechanism can generate and execute in parallel three consistency checks:

\begin{enumerate}
    \item $C_1 = \{ c_1,c_2 \}$ - the consistency check, which is being required by \textsc{FD}.

    \item $C_2 = \{ c_1 \}$ - the first half of $C$, which will be checked in the next \textsc{FD} call if $\{ c_1,c_2 \}$ is inconsistent.
    
    \item $C_3 = \{ c_1,c_2,c_3 \}$ - a union of $C$ and the first half of $\rho$, which will be checked if $\{ c_1,c_2 \}$ is consistent.
    
\end{enumerate}

Since the \emph{look-ahead} mechanism runs on one CPU core, only three future consistency checks are generated in our example. Each generated consistency check is asynchronously executed in one core.


\textbf{\textsc{LookUp} table.} Consistency checks generated by the \emph{look-ahead} mechanism are stored in a global \textsc{LookUp} table 
(see \textit{Table \ref{tab:Lookup-table}}). If \textsc{FD} needs to know the consistency of a given set of constraints, a simple lookup is triggered to get the corresponding consistency check's result. Assume that there is no consistency check for the requested set in the \textsc{LookUp} table. In that case, the algorithm runs the \emph{look-ahead} mechanism to generate and execute in parallel anticipated consistency checks. Consistency checks in the \textsc{LookUp} table can also be exploited to restrict the generation of the consistency checks that have already been created in the previous steps of the \emph{look-ahead} mechanism. This way, all anticipated consistency checks can be done only once and will not waste computer resources.

\begin{table}[t]
\small
\centering 
\begin{tabular}{|l|l|c|c|c|}
\hline
node-id & constraint set 	& consistent   
\tabularnewline
\hline
$1$ & $\{C_{KB} \cup \{c_{10} \twodots c_{13}\}\}$ & $false$
\tabularnewline
\hline
$1.2$ & $\{C_{KB} \cup \{c_{10},c_{11}\}\}$ & $false$
\tabularnewline
\hline
$1.2.1$ & $\{C_{KB} \cup \{c_{10},c_{11},c_{12}\}\}$ & --
\tabularnewline
\hline
$1.2.1.2$ & $\{C_{KB} \cup \{c_{10},c_{11},c_{13}\}\}$ & --
\tabularnewline
\hline
$1.2.2$ & $\{C_{KB} \cup \{c_{10}\}\}$ & $true$
\tabularnewline
\hline
\end{tabular}
\caption{A \textsc{LookUp} table created in our working example, including part of consistency checks generated by \textsc{AddCC} in the \textsc{LookAhead} function (see \textit{Figure \ref{fig:fastdiagp-ex-tree}}) and executed in parallel. The `--' entries indicate that the corresponding consistency checks are still ongoing or have not been started.} 
\label{tab:Lookup-table}
\end{table}

\textbf{\textsc{Consistent} function.} \textsc{FastDiagP} uses the \textsc{Consistent} function (see \textit{Algorithm \ref{alg:isconsistent}}) that requires three inputs: a consideration set $C$, a background knowledge $B$, and a set of constraints $\rho$ that has not been checked yet. Different from the \textsc{Consistent} function in \textsc{FastDiag}, the additional parameter $\rho$ is needed to help the \emph{look-ahead} mechanism conduct inferences about future needed consistency checks. Since these sets are \textsc{FD}'s inputs at each recursive step, no additional computations are required.

The \textsc{Consistent} function checks the existence of a consistency check for $B \cup C$ in the \textsc{LookUp} table. If this is the case,  the function returns the consistency check's outcome. Otherwise, it activates the \textsc{LookAhead} function (see \textit{Algorithm \ref{alg:lookahead}}) to generate further consistency checks that might be relevant in upcoming \textsc{FD} recursive calls.

\begin{algorithm}[t]
\caption{\textsc{Consistent}$(C, B, \rho): Boolean$}
\small
\begin{algorithmic}[1]\label{alg:isconsistent}
\IF{$\neg \textsc{ExistCC}(B \cup C)$}
\STATE \textsc{LookAhead}$(C, B, \{ \rho \})$
\ENDIF
\STATE \emph{return} \textsc{LookUp}$(B \cup C)$
\end{algorithmic}
\end{algorithm}

\textbf{\textsc{LookAhead} function.} The \emph{look-ahead} mechanism is implemented in the recursive \textsc{LookAhead} function (\textit{Algorithm \ref{alg:lookahead}}), requiring three parameters: (1) a consideration set $C$, (2) a background set $B$ holding the already-considered and \emph{assumed} consistent constraints, and (3) an ordered set $\phi$ in which each item is a set of constraints to be considered when $C$ is a singleton or assumed to be consistent.

The first constraint set $\phi_1$ of $\phi$ is always the most recent second subset divided from $C$ in the last recursive call. $\phi_1$ has to be considered first when \textsc{LookAhead} takes into account sets of $\phi$. With the structure of $\phi$, the order of consistency checks generated by \textsc{LookAhead} matches the order of consistency checks requested by \textsc{FastDiag}. For instance, in node [1.2.2] of the \textsc{LookAhead} execution trace (see \textit{Figure \ref{fig:fastdiagp-ex-tree}}), $\phi$ contains two sets: $\phi_1=\{ c_{11} \}$ and $\phi_2=\{ c_{12},c_{13} \}$. The set $\phi_2=\{ c_{12},c_{13} \}$, which is added to $\phi$ in node [1], is the second half of $C=\{ c_{10} \twodots c_{13} \}$. The set $\phi_1=\{ c_{11} \}$ separated from $C=\{ c_{10},c_{11} \}$ is added to $\phi$ in node [1.2]. Since $\phi_1$ is added later on, it will be considered first in the next \textsc{LookAhead} call. In particular, it is considered in node [1.2.2.2] before taking into account $\phi_2$ in node [1.2.2.2.1]. This mechanism works in a similar fashion as \textsc{FastDiag}. 


\begin{figure*}[hbt!]
\centerline{\includegraphics[scale=0.69]{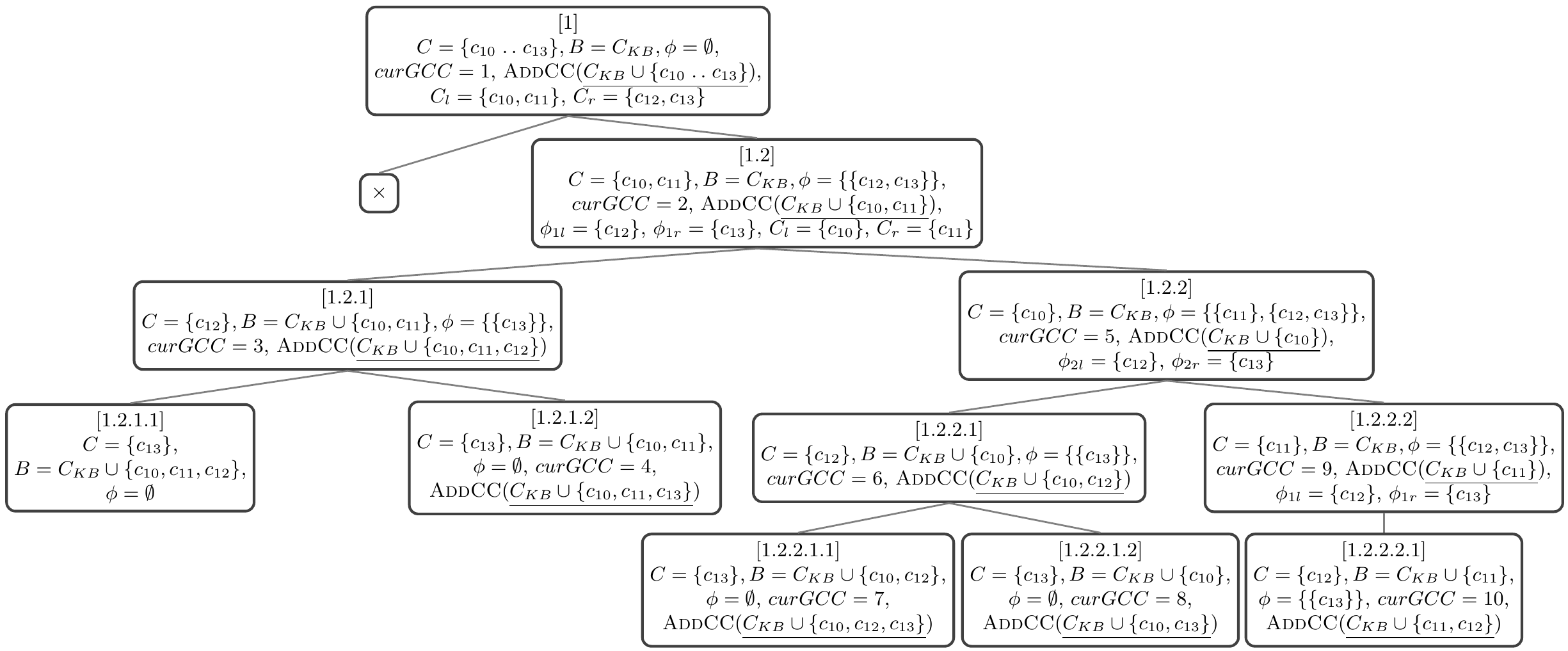}}
\caption{\textsc{LookAhead} execution trace for $C=\{ c_{10} \twodots c_{13} \}$, $B=C_{KB}$, $\phi = \emptyset$, and $maxGCC=10$. The consistency checks $\{C_{KB} \cup \{ c_{10},c_{11} \}\}$, $\{C_{KB} \cup \{ c_{10} \}\}$, $\{C_{KB} \cup \{ c_{10},c_{12} \}\}$, and $\{C_{KB} \cup \{ c_{10},c_{12},c_{13} \}\}$ generated by \textsc{LookAhead}, can be used by the \textsc{FastDiag} execution trace illustrated in \textit{Figure \ref{fig:fd-ex-tree}}.}
\label{fig:fastdiagp-ex-tree} 
\end{figure*}

\begin{algorithm}[t]
\caption{\\\textsc{LookAhead}$(C, B, \phi= \{\{\phi_1\} \twodots \{\phi_p\}\})$}
\small
\begin{algorithmic}[1]\label{alg:lookahead}
\IF{$curGCC < maxGCC$}
    \IF{$\neg \textsc{ExistCC}(B \cup C)$}
        \STATE $curGCC \leftarrow curGCC + 1$
        \STATE \textsc{AddCC}($B \cup C$)
    \ENDIF
    
    \STATE \{$B \cup C$ \textit{assumed consistent}\}
    \IF[case 1.1]{$|\phi| > 1 \land |\phi_1| = 1 \land \textsc{ExistCC}(B \cup C \cup \phi_1)$}
        \STATE \textsc{Split}$(\phi_2, \phi_{2l}, \phi_{2r})$
        \STATE \textsc{LookAhead}$(\phi_{2l}, B \cup C, \phi_{2r} \cup  (\phi \setminus \{ \phi_1, \phi_2 \}))$
    \ELSIF[case 1.2]{$\phi \neq \emptyset \land |\phi_1| = 1$}
        \STATE \textsc{LookAhead}$(\phi_1, B \cup C, \phi \setminus \phi_1)$
    \ELSIF[case 1.3]{$\phi \neq \emptyset \land |\phi_1| > 1$}
        \STATE \textsc{Split}$(\phi_1, \phi_{1l}, \phi_{1r})$
        \STATE \textsc{LookAhead}$(\phi_{1l}, B \cup C, \phi_{1r} \cup  (\phi \setminus \phi_1))$
    \ENDIF
    
    \STATE \{$B \cup C$ \textit{assumed inconsistent}\}
    \IF[case 2.1]{$|C| > 1$}
        \STATE \textsc{Split}$(C, C_l, C_r)$
        \STATE \textsc{LookAhead}$(C_l, B, C_r \cup \phi)$
    \ELSIF[case 2.2]{$|C| = 1 \land |\phi_1| = 1$}
        \STATE \textsc{LookAhead}$(\phi_1, B, \phi \setminus \phi_1)$
    \ELSIF[case 2.3]{$|C| = 1 \land |\phi_1| > 1$}
        \STATE \textsc{Split}$(\phi_1, \phi_{1l}, \phi_{1r})$
        \STATE \textsc{LookAhead}$(\phi_{1l}, B, \phi_{1r} \cup  (\phi \setminus \phi_1))$
    \ENDIF
\ENDIF
\end{algorithmic}
\end{algorithm}

In each \textsc{LookAhead} call, two global parameters $curGCC$ (initialized by zero) and $maxGCC$ (initialized by $\#cores$) are used to restrict the maximum number of generated speculative consistency checks. \textsc{LookAhead} checks the available space for further consistency checks and examines if any consistency check exists for $B \cup C$ so far. If not, then the $AddCC$ function is called to activate a consistency check for $B \cup C$ asynchronously and adds an entry of $B \cup C$ to the \textsc{LookUp} table. Next, \textsc{LookAhead} predicts potentially relevant consistency checks needed by \textsc{FD} based on \textit{two assumptions}: (1) \textit{$B \cup C$ is consistent} and (2) \textit{$B \cup C$ is inconsistent} (see the details in the next paragraphs). The order of two assumptions is opposite to this in \textsc{FastDiag} since consistency checks with larger cardinality should be executed in advance, which helps reduce the waiting time in case the corresponding consistency checks are still ongoing. 

\textbf{\textit{Assumption 1 ($B \cup C$ is consistent)}}: The function further checks the sets of $\phi$ when $B \cup C$ is consistent, i.e., all \textsc{LookAhead} calls will have $B \cup C$ as the background set.

\begin{itemize}
    \item If there is a consistency check for $B \cup C \cup \phi_1$ in the \textsc{LookUp} table (see \emph{case 1.1} in \textit{Algorithm \ref{alg:lookahead}}), \textsc{LookAhead} omits $\phi_1$ and further checks $\phi_2$. The function considers the first half of $\phi_2$ ($\phi_{2l}$) as the consideration set. $\{ \phi_1,\phi_2 \}$ in $\phi$ are replaced with the second half of $\phi_2$ (i.e., $\phi_{2r} \cup  (\phi \setminus \{ \phi_1, \phi_2 \}$)). Let's have a look at an example in \textit{Figure \ref{fig:fastdiagp-ex-tree}}. In node [1.2.2], the consistency check for $C_{KB} \cup \{ c_{10},c_{11} \}$ has already been generated. Hence, in node [1.2.2.1], the function omits $\phi_1=\{ c_{11} \}$ and proceeds with a further look ahead for $\phi_2$.
            
    \item If $|\phi_1| = 1$ (see \emph{case 1.2} in \textit{Algorithm \ref{alg:lookahead}}), the input parameters of the function are $\phi_1$, $B \cup C$, and $\phi \setminus \phi_1$. In our example  (\textit{Figure \ref{fig:fastdiagp-ex-tree}}), this case will be applied to expand the \emph{consistent} branch of node [1.2.2.1].
            
    \item If $|\phi_1| > 1$ (see \emph{case 1.3} in \textit{Algorithm \ref{alg:lookahead}}), $\phi_1$ is divided into two halves $\phi_{1l}$ and $\phi_{1r}$. \textsc{LookAhead} is called where $C$ is replaced with the first half $\phi_{1l}$ and $\phi_1$ in $\phi$ is replaced with the second half $\phi_{1r}$ (i.e., $\phi_{1r} \cup (\phi \setminus \phi_1)$). One example is shown in node [1.2] of \textit{Figure \ref{fig:fastdiagp-ex-tree}}.
\end{itemize}

\textbf{\textit{Assumption 2 ($B \cup C$ is inconsistent)}}: Consistency checks for the halves of $C$ are necessary to identify elements of $C$ responsible for the inconsistency.  
    
\begin{itemize}
    \item If the cardinality of $C$ is greater than 1 (see \emph{case 2.1} in \textit{Algorithm \ref{alg:lookahead}}), $C$ is divided into two halves $C_l$ and $C_r$. Thereafter, \textsc{LookAhead} for $C_l$ is called, where $B \cup C_l$ becomes the next consistency check, and $C_r$ is stored in $\phi$ to be considered when $B \cup C_l$ is assumed to be consistent or $C_l$ is a singleton. Note that $C_r$ is added to the head of $\phi$ to be considered first when the function takes into account the sets of $\phi$. Examples of this case are shown in nodes [1] and [1.2] of \textit{Figure \ref{fig:fastdiagp-ex-tree}}.
        
    \item If $|C| = 1$, the function further checks the sets of $\phi$ with the \emph{inconsistency} assumption: 
        \begin{itemize}
            \item If the first set of $\phi$ is a singleton ($|\phi_1| = 1$) (see \emph{case 2.2} in \textit{Algorithm \ref{alg:lookahead}}), \textsc{LookAhead} is called, where $C$ is replaced with $\phi_1$ and $\phi_1$ is removed from $\phi$. The execution of \textsc{LookAhead} according to this case can be found in nodes [1.2.1] and [1.2.2] of \textit{Figure \ref{fig:fastdiagp-ex-tree}}.
            
            \item If $\phi_1$ consists of several constraints to be considered ($|\phi_1| > 1$) (see \emph{case 2.3} in \textit{Algorithm \ref{alg:lookahead}}), $\phi_1$ is divided into two halves $\phi_{1l}$ and $\phi_{1r}$. The function \textsc{LookAhead} is called, where the first half $\phi_{1l}$ is the consideration set, $\phi_1$ in $\phi$ is replaced with $\phi_{1r}$ (i.e., $\phi_{1r} \cup (\phi \setminus \phi_1)$). In \textit{Figure \ref{fig:fastdiagp-ex-tree}}, the \emph{inconsistent} branch of node [1.2.2.2] is created based on this case.
        \end{itemize}
\end{itemize}


\section{Theoretical Analysis of \textsc{FastDiagP}}\label{sec:analysis}

\textbf{Soundness and Completeness of \textsc{FastDiagP}.} \textsc{FastDiagP} preserves the soundness and completeness properties of \textsc{FastDiag}. \textsc{FastDiagP} does not change the \textsc{FD} function but integrates the \emph{look-ahead} mechanism, where some consistency checks requested by \textsc{FD} are pre-calculated. Besides, \textsc{FD} performs simple lookups instead of expensive solver calls. Therefore, \textsc{FastDiagP} obtains better performance, and the returned diagnosis is minimal and preferred.

\textbf{Soundness and Completeness of \textsc{LookAhead}.} \textsc{LookAhead} generates correct consistency checks for subsets of the consideration set $C$. Assuming that $C'$ is a set generated by \textsc{LookAhead}, but $C' \not \subseteq (C \cup \phi)$. We can say that $\exists c_k: c_k \in C' \land c_k \not \in (C \cup \phi)$. Since \textsc{LookAhead} does not exploit constraints outside of $C \cup \phi$, $C'$ is not a set generated by \textsc{LookAhead}. Besides, by following two principles \emph{P1} and \emph{P2}, \textsc{LookAhead} can generate all possible combinations of constraints in $C$.

\textbf{Uniqueness of anticipated consistency checks.} \textsc{LookAhead} exploits $\phi$ in both assumption branches, which leads to redundant consistency checks when exploiting $\phi$ according to the \emph{consistency} assumption. The checks in lines 2 and 7 (\textit{Algorithm \ref{alg:lookahead}}) assure that generated consistency checks are unique.

\textbf{Complexity analysis.} \emph{\textsc{FD} complexity}. The \emph{worst-case} complexity of \textsc{FD} in terms of the number of needed consistency checks for determining MSS $\Omega$ and the corresponding diagnosis $\Delta = C - \Omega$ is $2d \times log_2(\frac{n}{d})+2d$, where $d$ is the \emph{set size of the minimal diagnosis}, $n$ is the \emph{number of constraints} in $C$, and $2d$ represents the branching factor and the number of leaf-node consistency checks. The \emph{best-case} complexity is $log_2(\frac{n}{d})+2d$. In the worst case, each diagnosis element is located in a different path of the search tree. The factor $log_2(\frac{n}{d})$ represents the depth of a path of the \textsc{FD} search tree. In the best case, all constraints part of a diagnosis are included in a single path of the search tree.

\emph{\textsc{LookAhead} complexity}. Assuming that $\phi=\emptyset$, the number of consistency checks ($N$) generated by \textsc{LookAhead} is the sum of all possible combinations of $n$ constraints in the consideration set $C$. It means that $N=\sum_{i=1}^{n}{{n}\choose{i}} = 2^{n}-1$. Due to the uniqueness of \textsc{LookAhead}, the upper bound of its space complexity in terms of the number of \textsc{LookAhead} calls is $2^{n}-1$.

\textbf{Termination of \textsc{LookAhead}}. If $maxGCC \leq 2^{n}-1$, recursive calls of \textsc{LookAhead} stop when $maxGCC$ consistency checks are generated. Otherwise, \textsc{LookAhead} terminates if $C$ and $\phi$ are empty.

\section{Empirical Evaluation}\label{sec:evaluation}


\textbf{Experiment design.} In this study, we compared the performance of \textsc{FastDiagP} and \textsc{FastDiag} according to three aspects: (1) \textit{run-time} $R$ needed to determine \textit{the preferred diagnosis}, (2) \emph{speedup} $S$ that tells us the gain we get through the parallelization, and (3) \emph{efficiency} $E$ representing the ratio between the speedup and the number of processes in which we run the algorithm. In particular, \emph{speedup} $S_p$ is computed as $=T_1/T_p$, where $T_1$ is the wall time when using 1 core (\textsc{FastDiag}) and $T_p$ is the wall time when $p$ cores are used. The efficiency $E_p$ is defined as $S_p/p$. These aspects were analyzed in two dimensions: the \textit{diagnosis cardinality} and the \textit{available computing cores} ($\#cores$).

\textbf{Dataset and Procedure.} The basis for these evaluations was the \emph{Linux-2.6.33.3} configuration knowledge base taken from Diverso Lab's benchmark\footnote{https://github.com/diverso-lab/benchmarking} \cite{Heradio2022}. The characteristics of this knowledge base are the following: $\#$features = 6,467; $\#$relationships = 6,322; and $\#$cross-tree constraints = 7,650. For this knowledge base, we randomly synthesized\footnote{To ensure the reproducibility of the results, we used the seed value of 141982L for the random number generator.} and collected 20,976 inconsistent sets of requirements, whose cardinality ranges from 5 to 250. 
We applied systematic sampling technique \cite{mostafa2018} to select 10 inconsistent requirements with diagnosis cardinalities of 1, 2, 4, 8, and 16.



The diagnosis algorithms were implemented in \textit{Python} using \textsc{Sat4j} \cite{leberre2010sat4j} as a reasoning solver.\footnote{The dataset, the implementation of algorithms, and evaluation programs can be found at \emph{https://github.com/AIG-ist-tugraz/FastDiagP}.} We used the \textsc{CNF} class of \textsc{PySAT} \cite{pysat18} for representing constraints and the \textit{Python multiprocessing} package for running parallel tasks. All experiments reported in the paper were conducted with an Amazon EC2 instance\footnote{https://aws.amazon.com/ec2/instance-types/c5/} of the type c5a.8xlarge, offering 32 vCPUs with 64-GB RAM.


\begin{table}[hbt!]
\small
\centering 
\begin{tabular}{|c|c|c|c|c|c|c|c|}
\hline
 	\multicolumn{3}{|c|}{} & \multicolumn{5}{c|}{$\#cores$}\\
 	\cline{4-8}
 	\multicolumn{3}{|c|}{} & \centering 1 & \centering 4 & \centering 8 & \centering 16 & 32 \\
 	\hline
 	\multirow{15}{*}{$|diag|$} & \multirow{3}{*}{1} &
 	    $R$ & 4.56 & \textbf{3.08} & \textbf{2.77} & \textbf{2.63} & \textbf{3.29} \\
        \cline{3-8}
        & & $S$ & & 1.48 & 1.65 & 1.74 & 1.39 \\
        \cline{3-8}
        & & $E$ & & 0.49 & 0.24 & 0.12 & 0.05 \\
        \cline{2-8}
    & \multirow{3}{*}{2} &
 	    $R$ & 5.60 & \textbf{4.00} & \textbf{3.69} & \textbf{3.71} & \textbf{5.05} \\
        \cline{3-8}
        & & $S$ & & 1.40 & 1.52 & 1.51 & 1.11 \\
        \cline{3-8}
        & & $E$ & & 0.47 & 0.22 & 0.10 & 0.04 \\
        \cline{2-8}
    & \multirow{3}{*}{4} &
 	    $R$ & 8.13 & \textbf{5.95} & \textbf{5.76} & \textbf{6.43} & 10.11\\
        \cline{3-8}
        & & $S$ & & 1.37 & 1.41 & 1.26 & 0.80 \\
        \cline{3-8}
        & & $E$ & & 0.46 & 0.20 & 0.08 & 0.03 \\
        \cline{2-8}    
    & \multirow{3}{*}{8} &
 	    $R$ & 11.96 & \textbf{9.06} & \textbf{8.74} & \textbf{9.63} & 14.52\\
        \cline{3-8}
        & & $S$ & & 1.32 & 1.37 & 1.24 & 0.82 \\
        \cline{3-8}
        & & $E$ & & 0.44 & 0.20 & 0.08 & 0.03 \\
        \cline{2-8}
    & \multirow{3}{*}{16} &
 	    $R$ & 20.95 & \textbf{16.80} & \textbf{16.38} & \textbf{19.02} & 29.28\\
        \cline{3-8}
        & & $S$ & & 1.25 & 1.28 & 1.10 & 0.72 \\
        \cline{3-8}
        & & $E$ & & 0.42 & 0.18 & 0.07 & 0.02 \\
        \cline{2-8}
\hline
\end{tabular}
\caption{\textit{Average runtime} $R$ (in $sec$), \textit{speedup} $S$, and \textit{efficiency} $E$ of \textsc{FastDiagP} ($\#cores>1$) versus \textsc{FastDiag} ($\#cores=1$) needed for determining the preferred diagnosis with a repetition rate of 3 per setting and $maxGCC=\#cores - 1$. $|diag|$ denotes the cardinality of the diagnosis.}
\label{tab:pFDresults3}
\end{table}

\begin{table}[hbt!]
\small
\centering 
\begin{tabular}{|c|c|c|c|c|c|}
\hline
 	\multicolumn{2}{|c|}{} & \multicolumn{4}{c|}{$\#cores$}\\
 	\cline{3-6}
 	\multicolumn{2}{|c|}{} & \centering 1 & \centering 8 & \centering 16 & 32 \\
 	\hline
 	\multirow{5}{*}{$|diag|$} & 1 & 4.56 & \textbf{2.78} & 2.79 & 2.81 \\
    \cline{2-6}
    & 2 & 5.60 & \textbf{3.69} & 3.70 & 3.72\\
    \cline{2-6}
    & 4 & 8.13 & \textbf{5.76} & 5.80 & 5.87\\
    \cline{2-6}
    & 8 & 11.96 & \textbf{8.74} & 8.78 & 8.84\\
    \cline{2-6}
    & 16 & 20.95 & \textbf{16.38} & 16.41 & 16.64\\
\hline
\end{tabular} 
\caption{\textit{Average runtime} (in $sec$) of \textsc{FastDiagP} ($\#cores>1$) versus \textsc{FastDiag} ($\#cores=1$) needed for determining the preferred diagnosis with a repetition rate of 3 per setting and $maxGCC=7$. $|diag|$ denotes the cardinality of the preferred diagnosis. The \textit{bold values} prove the optimal number of CPU cores ($\#cores=8$).}
\label{tab:pFDresults4}
\end{table}

\textbf{Results.} The experimental results show that \textsc{FastDiagP} outperforms the sequential direct diagnosis approach in almost all scenarios (see the \textbf{bold values} in\textit{ Table \ref{tab:pFDresults3}}). Besides, in \textit{Tables \ref{tab:pFDresults3}} and \textit{\ref{tab:pFDresults4}}, the optimal number of CPU cores is 8 and the corresponding speedup values range from 1.28 to 1.65, showing the runtime deduction up to 40\%. The $\#cores$ higher than 8 becomes less efficient for boosting the performance. Particularly, in \textit{Table \ref{tab:pFDresults3}}, 
the increase of $\#cores$ (that also triggers the increase of $maxGCC$ ($maxGCC = \#cores - 1 $)) leads to gradual runtime increase (i.e., lower performance). A parallelization mechanism with more than 8 cores is not so much helpful in such a scenario. This manifests when $\#cores=32$, $maxGCC=31$, and $|diag|=4,8,16$. The reason is that the \textsc{LookAhead} function applies a sequential mechanism. When $maxGCC$ gets higher, the runtime of \textsc{LookAhead} increases exponentially, leading to a significant increase of \textsc{FastDiagP}'s runtime. Besides, \textit{Table \ref{tab:pFDresults4}} confirms that the utilization of more than 8 CPU cores becomes less efficient. In this evaluation, our idea was to fix the $maxGCC$ value ($maxGCC=7$) to see how the performance of \textsc{FastDiagP} is when $\#cores$ is higher than 8.






\section{Conclusion}\label{sec:conclution}
In this paper, we have proposed a parallelized variant of the \textsc{FastDiag} algorithm to diagnose over-constrained problems. Our parallelized approach helps to exploit multi-core architectures and provides an efficient preferred diagnosis detection mechanism. Furthermore, our approach is helpful for dealing with complex over-constrained problems, boosting the performance of various knowledge-based applications, and making these systems more accessible, especially in the context of interactive settings. Open topics for future research are the following: (1) performing more in-depth evaluations on the basis of industrial configuration knowledge bases (in this context, we plan to analyze the different \emph{look-ahead} search approaches, e.g. breadth-first search, in further detail), and (2) applying speculative reasoning for supporting anytime diagnosis tasks.




\section*{Acknowledgements}

This work has been partially funded by the FFG-funded project \textsc{ParXCel} (880657) and two other projects \textsc{Copernica} (P20\_01224) and \textsc{MetaMorFosis} (FEDER\_US-1381375) funded by Junta de Andaluc\'{i}a.

\bibliography{references.bib}

\begin{thebibliography}{27}
\providecommand{\natexlab}[1]{#1}

\bibitem[{Benavides, Segura, and Ruiz-Cort{\'e}s(2010)}]{Benavides2010}
Benavides, D.; Segura, S.; and Ruiz-Cort{\'e}s, A. 2010.
\newblock Automated Analysis of Feature Models 20 Years Later: A Literature
  Review.
\newblock \emph{Information Systems}, 35(6): 615--636.

\bibitem[{Bordeaux, Hamadi, and Samulowitz(2009)}]{Bordeauxetal2009}
Bordeaux, L.; Hamadi, Y.; and Samulowitz, H. 2009.
\newblock {Experiments with Massively Parallel Constraint Solving}.
\newblock In \emph{21st International Joint Conference on Artifical
  Intelligence}, 443--448. California, USA: Morgan Kaufmann.

\bibitem[{{Burton}(1985)}]{Burton1985}
{Burton}, F.~W. 1985.
\newblock Speculative computation, parallelism, and functional programming.
\newblock \emph{IEEE Transactions on Computers}, C-34(12): 1190--1193.

\bibitem[{Castillo et~al.(2005)Castillo, Borrajo, Salido, and
  Oddi}]{Castillo2005}
Castillo, L.; Borrajo, D.; Salido, M.; and Oddi, A. 2005.
\newblock \emph{Planning, Scheduling and Constraint Satisfaction: From Theory
  to Practice}, volume 117 of Frontiers in Artificial Intelligence and
  Applications.
\newblock IOPress.

\bibitem[{Felfernig and Burke(2008)}]{FelfernigBurke2008}
Felfernig, A.; and Burke, R. 2008.
\newblock {Constraint-based Recommender Systems: Technologies and Research
  Issues}.
\newblock In \emph{ACM International Conference on Electronic Commerce
  (ICEC'08)}, 17--26. Innsbruck, Austria.

\bibitem[{Felfernig et~al.(2004)Felfernig, Friedrich, Jannach, and
  Stumptner}]{Felfernig2014_Elsevier}
Felfernig, A.; Friedrich, G.; Jannach, D.; and Stumptner, M. 2004.
\newblock Consistency-based diagnosis of configuration knowledge bases.
\newblock \emph{Artificial Intelligence}, 152: 213--234.

\bibitem[{Felfernig et~al.(2009)Felfernig, Friedrich, Schubert, Mandl,
  Mairitsch, and Teppan}]{felfernig2009plausible}
Felfernig, A.; Friedrich, G.; Schubert, M.; Mandl, M.; Mairitsch, M.; and
  Teppan, E. 2009.
\newblock Plausible Repairs for Inconsistent Requirements.
\newblock In \emph{Proceedings of the 21st International Jont Conference on
  Artifical Intelligence (IJCAI'09)}, 791–796. California, USA: Morgan
  Kaufmann.

\bibitem[{Felfernig et~al.(2014)Felfernig, Hotz, Bagley, and
  Tiihonen}]{Felfernig2014}
Felfernig, A.; Hotz, L.; Bagley, C.; and Tiihonen, J. 2014.
\newblock \emph{Knowledge-based Configuration: From Research to Business
  Cases}.
\newblock San Francisco, CA, USA: Morgan Kaufmann Publishers Inc., 1 edition.
\newblock ISBN 012415817X, 9780124158177.

\bibitem[{Felfernig et~al.(2010)Felfernig, Schubert, Mandl, Friedrich, and
  Teppan}]{Felfernig2010_ECAI}
Felfernig, A.; Schubert, M.; Mandl, M.; Friedrich, G.; and Teppan, E. 2010.
\newblock Efficient Explanations for Inconsistent Constraint Sets.
\newblock In \emph{{ECAI} 2010 - 19th European Conference on Artificial
  Intelligence, Lisbon, Portugal, August 16-20, 2010}, 1043--1044. {IOS} Press.

\bibitem[{Felfernig, Schubert, and Zehentner(2012)}]{Felfernig2012}
Felfernig, A.; Schubert, M.; and Zehentner, C. 2012.
\newblock An Efficient Diagnosis Algorithm for Inconsistent Constraint Sets.
\newblock \emph{Artif. Intell. Eng. Des. Anal. Manuf.}, 26(1): 53--62.

\bibitem[{Felfernig et~al.(2018)Felfernig, Walter, Galindo, Benavides, Erdeniz,
  Atas, and Reiterer}]{Felfernig2018}
Felfernig, A.; Walter, R.; Galindo, J.~A.; Benavides, D.; Erdeniz, S.~P.; Atas,
  M.; and Reiterer, S. 2018.
\newblock Anytime diagnosis for reconfiguration.
\newblock \emph{J. Intell. Inf. Syst.}, 51(1): 161--182.

\bibitem[{Gent et~al.(2018)Gent, Miguel, Nightingale, McCreesh, Prosser,
  Nooore, and Unsworth}]{Gentetal2018}
Gent, I.; Miguel, I.; Nightingale, P.; McCreesh, C.; Prosser, P.; Nooore, N.;
  and Unsworth, C. 2018.
\newblock {A Review of Literature on Parallel Constraint Solving}.
\newblock \emph{Theory and Practice of Logic Programming}, 18(5--6): 725--758.

\bibitem[{Heradio et~al.(2022)Heradio, Fernandez-Amoros, Galindo, Benavides,
  and Batory}]{Heradio2022}
Heradio, R.; Fernandez-Amoros, D.; Galindo, J.~A.; Benavides, D.; and Batory,
  D. 2022.
\newblock Uniform and scalable sampling of highly configurable systems.
\newblock \emph{Empirical Software Engineering}, 27(2): 44.

\bibitem[{Hotz et~al.(2014)Hotz, Felfernig, Stumptner, Ryabokon, Bagley, and
  Wolter}]{Hotz2014}
Hotz, L.; Felfernig, A.; Stumptner, M.; Ryabokon, A.; Bagley, C.; and Wolter,
  K. 2014.
\newblock {Configuration Knowledge Representation and Reasoning}.
\newblock In Felfernig, A.; Hotz, L.; Bagley, C.; and Tiihonen, J., eds.,
  \emph{Knowledge-based Configuration – From Research to Business Cases}, 41
  -- 72. Boston: Morgan Kaufmann.

\bibitem[{Ignatiev, Morgado, and Marques{-}Silva(2018)}]{pysat18}
Ignatiev, A.; Morgado, A.; and Marques{-}Silva, J. 2018.
\newblock {PySAT:} {A} {Python} Toolkit for Prototyping with {SAT} Oracles.
\newblock In \emph{SAT}, 428--437.

\bibitem[{Jannach, Schmitz, and Shchekotykhin(2015)}]{Jannachetal2015}
Jannach, D.; Schmitz, T.; and Shchekotykhin, K. 2015.
\newblock {Parallelized Hitting Set Computation for Model-Based Diagnosis}.
\newblock In \emph{29$^{th}$ AAAI Conference on Artificial Intelligence},
  1503--1510. Austin, Texas: AAAI Press.

\bibitem[{Jannach, Schmitz, and Shchekotykhin(2016)}]{Jannachetal2016}
Jannach, D.; Schmitz, T.; and Shchekotykhin, K. 2016.
\newblock {Parallel Model-Based Diagnosis on Multi-Core Computers}.
\newblock \emph{Journal of Artificial Intelligence Research}, 55: 835--887.

\bibitem[{Junker(2004)}]{junker2004quickxplain}
Junker, U. 2004.
\newblock {\textsc{QuickXPlain}: Preferred Explanations and Relaxations for
  over-Constrained Problems}.
\newblock In \emph{Proceedings of the 19th National Conference on Artifical
  Intelligence}, AAAI'04, 167–172. AAAI Press.

\bibitem[{Le et~al.(2021)Le, Felfernig, Uta, Benavides, Galindo, and
  Tran}]{le2021directdebug}
Le, V.-M.; Felfernig, A.; Uta, M.; Benavides, D.; Galindo, J.; and Tran, T.
  N.~T. 2021.
\newblock \textsc{DirectDebug}: Automated Testing and Debugging of Feature
  Models.
\newblock In \emph{2021 IEEE/ACM 43rd International Conference on Software
  Engineering: New Ideas and Emerging Results (ICSE-NIER)}, 81--85.

\bibitem[{Le~Berre and Parrain(2010)}]{leberre2010sat4j}
Le~Berre, D.; and Parrain, A. 2010.
\newblock The \textsc{Sat4j} library, release 2.2.
\newblock \emph{Journal on Satisfiability, Boolean Modeling and Computation},
  7(2-3): 59--64.

\bibitem[{Marques-Silva et~al.(2013)Marques-Silva, Heras, Janota, Previti, and
  Belov}]{MarquesSilva2013}
Marques-Silva, J.; Heras, F.; Janota, M.; Previti, A.; and Belov, A. 2013.
\newblock {On Computing Minimal Correction Subsets}.
\newblock In \emph{23rd International Joint Conference on Artificial
  Intelligence}, 615--622. Beijing, China.

\bibitem[{Marques-Silva and Previti(2014)}]{marquessilva2014preferred}
Marques-Silva, J.; and Previti, A. 2014.
\newblock On Computing Preferred MUSes and MCSes.
\newblock In \emph{Theory and Applications of Satisfiability Testing -- SAT
  2014}, 58--74. Cham: Springer.

\bibitem[{Mostafa and Ahmad(2018)}]{mostafa2018}
Mostafa, S.~A.; and Ahmad, I.~A. 2018.
\newblock Recent developments in systematic sampling: A review.
\newblock \emph{Journal of Statistical Theory and Practice}, 12(2): 290--310.

\bibitem[{Reiter(1987)}]{Reiter1987}
Reiter, R. 1987.
\newblock A Theory of Diagnosis from First Principles.
\newblock \emph{Artif. Intell.}, 32(1): 57--95.

\bibitem[{Rossi, van Beek, and Walsh(2006)}]{rossi2006handbook}
Rossi, F.; van Beek, P.; and Walsh, T. 2006.
\newblock \emph{Handbook of Constraint Programming}.
\newblock Elsevier.

\bibitem[{Stumptner(1997)}]{Stumptner1997}
Stumptner, M. 1997.
\newblock {An Overview of Knowledge‐based Configuration}.
\newblock \emph{Ai Communications}, 10(2): 111--125.

\bibitem[{Vidal et~al.(2021)Vidal, Felfernig, Galindo, Atas, and
  Benavides}]{Vidal2021}
Vidal, C.; Felfernig, A.; Galindo, J.; Atas, M.; and Benavides, D. 2021.
\newblock Explanations for over-constrained problems using \textsc{QuickXPlain}
  with speculative executions.
\newblock \emph{Journal of Intelligent Information Systems}, 57(3): 491--508.

\end{thebibliography}

\end{document}